\documentclass[conference]{IEEEtran}
\usepackage[hyphens,spaces,obeyspaces]{url}
\usepackage{graphicx}
\usepackage[backend=bibtex,style=numeric,natbib=true,sorting=none,giveninits=true,maxbibnames=9]{biblatex}
\usepackage[inline, shortlabels]{enumitem}
\usepackage{algorithm}
\usepackage{amsmath,bm}
\usepackage{algorithmic}
\usepackage{lipsum}
\usepackage[percent]{overpic}
\usepackage[T1]{fontenc}
\usepackage{textcomp}
\usepackage{booktabs}
\usepackage{multirow}
\usepackage{etoolbox,siunitx}
\usepackage{hyperref}
\usepackage{tikz}

\makeatletter
 \let\old@ps@headings\ps@headings
 \let\old@ps@IEEEtitlepagestyle\ps@IEEEtitlepagestyle
 \def\confheader#1{%
 \def\ps@headings{%
 \old@ps@headings%
 \def\@oddhead{\strut\hfill#1\hfill\strut}%
 \def\@evenhead{\strut\hfill#1\hfill\strut}%
 }%
 \def\ps@IEEEtitlepagestyle{%
 \old@ps@IEEEtitlepagestyle%
 \def\@oddhead{\strut\hfill#1\hfill\strut}%
 \def\@evenhead{\strut\hfill#1\hfill\strut}%
 }%
 \ps@headings%
 }
\makeatother

\confheader{%
2021 International Conference on Indoor Positioning and Indoor Navigation (IPIN), 29 Nov. -- 2 Dec. 2021, Lloret de Mar, Spain}

\IEEEoverridecommandlockouts
\IEEEpubid{\makebox[\columnwidth]
{\hfill 978-1-6654-0402-0/21/\$31.00~\copyright~2021 \text{IEEE}}
\hspace{\columnsep}\makebox[\columnwidth]{ }}

\makeatletter
\newcommand{\linebreakand}{%
  \end{@IEEEauthorhalign}
  \hfill\mbox{}\par
  \mbox{}\hfill\begin{@IEEEauthorhalign}
}
\makeatother

\begin{document}

\title{\LARGE \bf
Pluto: Motion Detection for Navigation in a VR Headset
}

\author{
\IEEEauthorblockN{\normalsize{Dmitri Kovalenko$^1$}}
\IEEEauthorblockA{\normalsize{dmitri.a.kovalenko@gmail.com}}
\and
\IEEEauthorblockN{\normalsize{Artem Migukin$^1$}}
\IEEEauthorblockA{\normalsize{migukin.a@gmail.com}}
\and
\IEEEauthorblockN{\normalsize{Svetlana Ryabkova}}
\IEEEauthorblockA{\normalsize{s.ryabkova@samsung.com}}
\and
\IEEEauthorblockN{\normalsize{Vitaly Chernov$^1$}}
\IEEEauthorblockA{\normalsize{vitaliy.chernov@gmail.com}}
\linebreakand 
\IEEEauthorblockN{\textit{Samsung R\&D Institute Russia}}
\IEEEauthorblockA{}
}

\thispagestyle{empty}
\pagestyle{empty}
\maketitle

\begin{abstract}
Untethered, inside-out tracking is considered a new goalpost for virtual reality,
  which became attainable with advent of machine learning in SLAM.
Yet computer vision-based navigation is always at risk of a tracking failure due to poor illumination or saliency of the environment.
An extension for a navigation system is proposed,
  which recognizes agents \textit{motion} and \textit{stillness}
  states with 87\% accuracy from accelerometer data.
40\% reduction in navigation drift is demonstrated in
  a repeated tracking failure scenario on a challenging dataset.
\end{abstract}

\section{Introduction} \label{sec:intro}
The virtual reality has been a subject of extensive research for decades.
The advancements in algorithms and hardware brought commodity setups  to the mass-market.
Novel systems provide untethered experience, where a user is not constrained by the cable connecting a headset to a computer.
Spatial restrictions due to the infra-red beacon operation range,
  inherent in \textit{outside-in} navigation~\cite{quinones2018hive}, were removed with the \textit{inside-out} navigation approach,
  supported by SLAM~(Simultaneous Localization and Mapping) running onboard,
  processing stereo or depth camera images,
  enhanced by sensor fusion with accelerometer and gyroscope data
  ~\cite{jones2011visual}.

Inertial sensors providing an input to a navigation system enable high-frequency position updates,
  but are impractical as a sole data source, as the double integration error accumulates~\cite{verplaetse1996inertial}.
Yet the situation may arise when the camera-based navigation system enters a failure mode due to
 motion blur, textureless environments, low illumination, occlusions, and rigid world assumption violations.
Current headsets halt position updates until the tracking is reestablished.
The user experience would not have been disrupted if a navigation system was equipped with a fallback technique for position estimates during a tracking failure.
A proprioceptive nature of IMU~(Inertial Measurement Unit) makes it a valid candidate for sensor fusion in SLAM, as wealth of information may be inferred from its data with help of machine learning.
The proposed method recognizes \textit{stillness} and \textit{motion} states of an agent and sends linear acceleration and velocity pseudo-updates to a navigation system,
thus reducing positioning drift. We named it Pluto.
Main contributions are:\footnote{D.~Kovalenko, A.~Migukin and V.~Chernov made contributions during their time in
   Samsung R\&D Institute Russia, but currently are working for Yandex, Huawei and Align Technology respectively.}
\begin{enumerate*}[series = tobecont, itemjoin = \quad, label=\arabic*.]
    \item Novel motion detector neural network, based on Temporal Causal Convolutions~\cite{bai1803empirical};
    \item Nagivation system, robust to tracking failures, embodied in a prototype of virtual reality headset (Fig.~\ref{gr:pluto}.c);
    \item Evaluation on the real-world dataset of highly discontinuous movements (Fig.~\ref{gr:pluto}.d)
\end{enumerate*}.

\section{Related Work} \label{sec:rel_work}
Even though Pluto is a visual-inertial navigation system,
    the main focus of this inquiry is a tracking failure handling,
    when the system degenerates into a purely inertial one,
    which has been addressed by a great number of works.
The excellent survey~\cite{harle2013survey} on Pedestrian Dead Reckoning recognizes two different approaches:
\begin{enumerate*}[series = tobecont, itemjoin = \quad, label=\arabic*.]
    \item Inertial Navigation Systems;
    \item Step and Heading Systems
\end{enumerate*}.
The proposed method would fall into the former category under such a classification.

\subsection{Inertial Navigation Systems}
These systems utilize "zero velocity" updates (ZUPT) as a measure to contain the double integration error.
Technically, updates are velocity resets, triggered by a step event detector, analyzing accelerometer data for peaks,
    as shown in~\cite{xiaofang2014applications}.
The detector accuracy is higher when the inertial sensor registers acceleration, related solely to a gait, not other body motions.
Hence the sensor placement on a foot is widely preferred;
    Skog et al.~\cite{skog2010evaluation} take it to a limit, creating a low-drift system with a sensor box of IMU and pressure sensors, placed inside the heel of a boot.
The velocity trend, caused by integration of accelerometer bias, is removed in~\cite{feliz2009pedestrian}.
Gusenbauer et al.~\cite{gusenbauer2010self} consider extra correction sources:
    localization in a radio footprint map, plausibility check with a building plan.
Most works report $0.14-2.3\%$ relative positioning error,
the lowest achieved on level ground with a constant gait cycle, ensured by a metronome~\cite{skog2010evaluation},
while higher drift was registered on rough and sandy terrains by Sun~\cite{sun2014zupt}.
Above works mostly consider 2D navigation,
    because the gravity influence on acceleration measurements is stronger along the vertical axis.
An extension to 2.5D was achieved by the floor change recognition in~\cite{puyol2014pedestrian}.

Notable contributions made recently are: \cite{cortes2018deep}, \cite{feigl2020rnn},
  where a neural network was trained to infer 3D velocity of an agent,
      providing a useful constraint to EKF~(Extended Kalman Filter).
In \cite{silva2019end}, a convolutional LSTM~(Long Short-Term Memory) network
  was applied to public datasets in an end-to-end fashion, to replace EKF as a source of 6D inertial odometry.
Liu et al.~\cite{liu2020tlio} propose a neural network estimating 3D displacements along with their uncertainties from IMU data,
which enables to use displacements as a source within a probabilistic framework of EKF.

The evolution in ZUPT publications is three-dimensional:
  \begin{enumerate*}[series = tobecont, itemjoin = \quad, label=\arabic*.]
      \item Stronger neural networks~\cite{wagstaff2018lstm};
      \item Cheaper IMU sensors, setups with fewer IMUs;
      \item More natural placements~\cite{yan2018ridi}, head-mounted and hand-held, instead of belt strap-down or shoe mount
  \end{enumerate*}.
The present study is aligned with all three dimensions, utilizing a single low-cost head-mounted IMU for navigation.

\subsection{Step and Heading Systems}
The alternative approach is not to integrate acceleration at all,
  but to count steps and infer their direction.
The foot-mounted sensor boxes are not required in that case,
  as demonstrated with a belt strap-down in~\cite{goyal2011strap}.
Goyal et al. derive heading from 3D attitude,
  implying only a forward movement of an agent,
  which might be true for many commuter scenarios,
  but does not hold for a virtual reality interaction.
Jiang et al. estimate  heading as a direction of spectrally filtered accelerometer data~\cite{jiang2015human}.
A neural network provided an online step length calibration in~\cite{beauregard2006helmet} for a mixed indoor/outdoor case,
  where GPS was exploited.
Beauregard et al. were among a few, who also considered a head-mounted setup.
Some pedestrian dead reckoning systems rely on an assumption that an agent moves only forward,
  having its orientation and heading always aligned.
Windau et al. removes this constraint in their work~\cite{windau2016walking}.

The inherent discretisation of the step and heading approach makes it less enticing for virtual reality,
  hence we forgo that in the Pluto navigation system.

\subsection{Action Classification}
A coalescent area of research, not being directly involved with navigation, is highly relevant: human action classification.
Attempts to recognize pedestrian action were conducted with various setups: from a minimalistic wristband~\cite{gjoreski2015recognizing} to a set of 19 IMUs mounted on every part of a human body~\cite{ordonez2016deep}, with 95\% $F_1$ score reported by latter with Deep Convolutional LSTM for 17 actions classification.
Anguita et al.~\cite{anguita2012human} propose a waistband dataset and SVM action classifier performing with $89\%$ accuracy.
This inquiry was taken further with Recurrent Neural Networks, achieving $94\%$ on the same dataset in~\cite{chevalier}.

3-class time series classifier was implemented~\cite{sun2008activity},
  discerning a steady walk,
  rest and irregular motions.
Contrary to the above works,
  Sun et al.~\cite{sun2008activity} have also demonstrated a way the classifier influences the navigation system,
  halting step counting during two latter states.
A similar classifier was put forward more recently in~\cite{rantanen2018}.

The present study went beyond accuracy evaluation,
  investigating detection delays and intervals between false positives,
  which provides insights into the system behavior over time.
We proceed with detailing the design of navigation system and motion detector network.

\section{Pluto Navigation System} \label{sec:overview}
\begin{figure}[t!]
\centering



\begin{tabular}{ccc}
  \multirow{3}*[4.6em]{\includegraphics[height=.23\textheight]{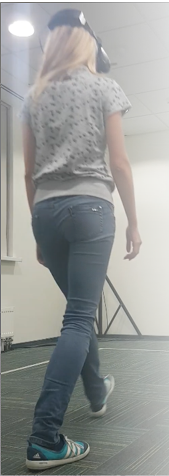}}
     & \includegraphics[height=.08\textheight]{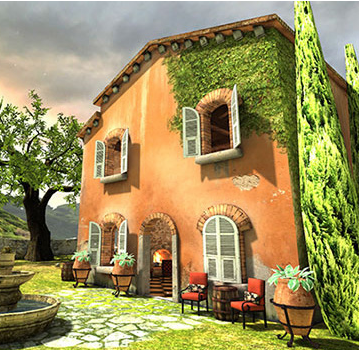} &  \includegraphics[height=.08\textheight]{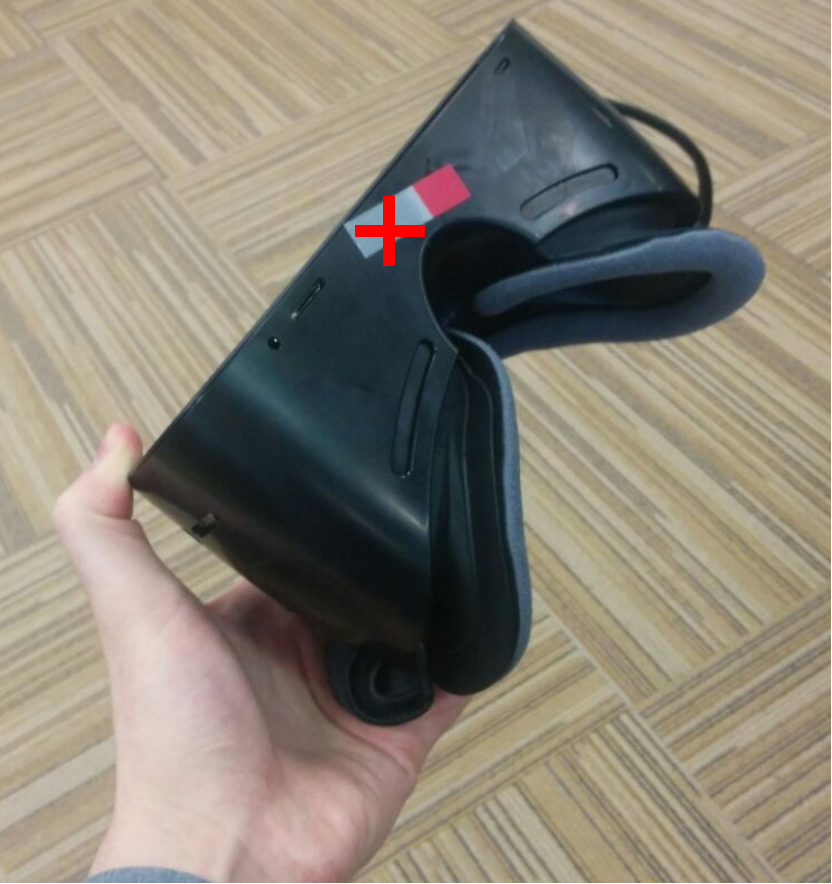} \\
     & (b) & (c) \\
     & \multicolumn{2}{c}{\includegraphics[height=.14\textheight]{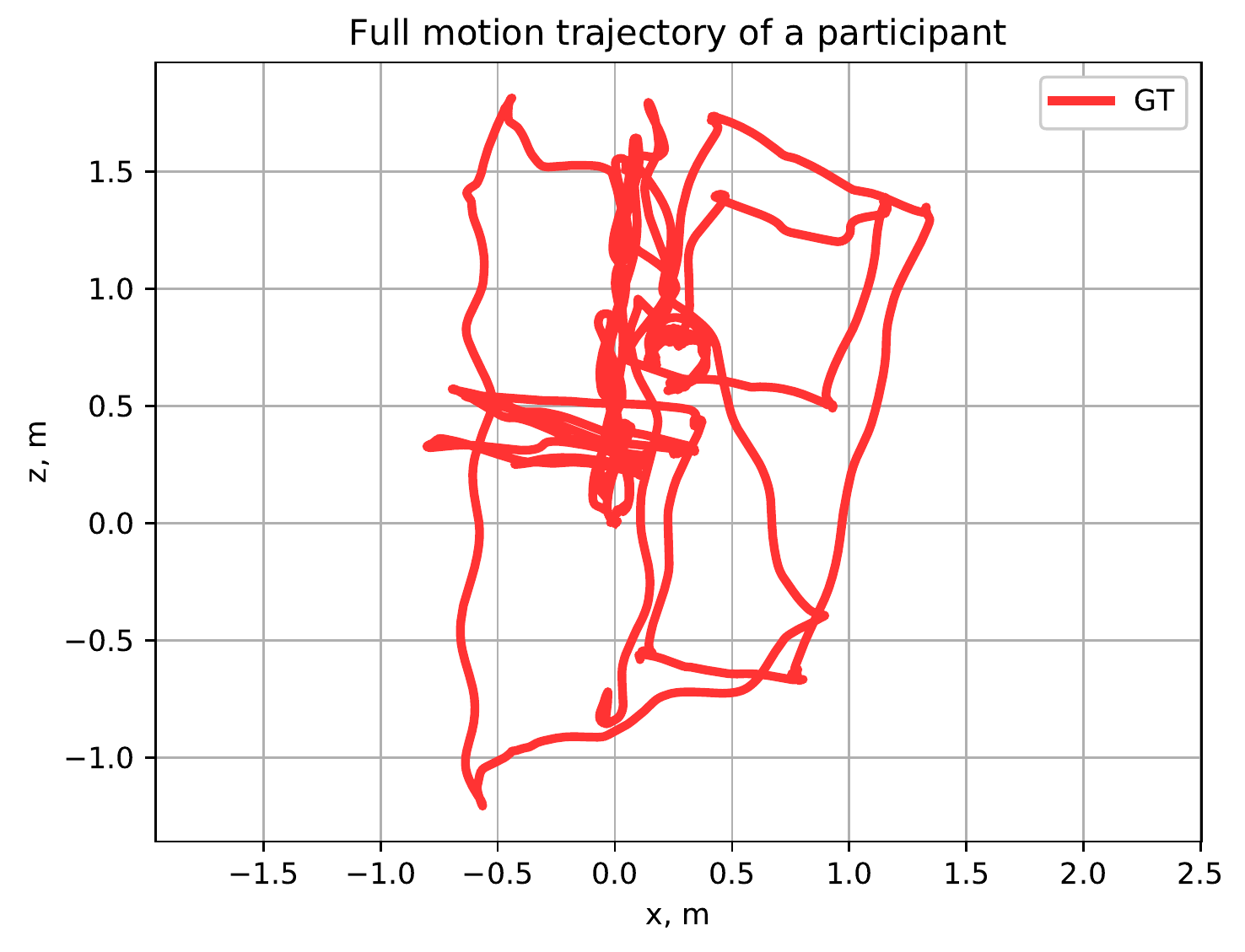}} \\
   (a) & \multicolumn{2}{c}{(d)} \\
\end{tabular}

\caption{
 a) An experiment participant taking a backward step
 b) The virtual scene explored by study participants
 c) The virtual reality headset prototype; the IMU sensor placement is marked (approximately) by a cross
 d) A participant movement trajectory in a bird-eye view during the study;
  no instructions on a gait pattern were given, except to explore a virtual reality scene naturally
 }
\label{gr:pluto}
\end{figure}

The headset prototype (Fig.~\ref{gr:pluto}.c),
    implementing virtual reality capabilities and hosting the online visual-inertial navigation system~(Pluto) was devised.
The orientation estimation system is purely inertial and closely follows~\cite{lavalle2014head},
 using complementary filters with accelerometer and magnetometer data for the tilt and yaw drift correction.
The magnetic field is modeled as in~\cite{ozyagcilar2012calibrating},
    with an additional capability to detect magnetic field anomalies and recalibrate when necessary.
Strong correlations were registered between IMU temperature and magnitudes of accelerometer and gyroscope noise.
Instead of random walk bias modeling, the online calibration accumulates samples and solves for biases
    for every temperature value~(IMU provides temperature measurements at $0.5^{\circ}$ C increments).
Optimal sample sizes are determined accordingly with~\cite{foi2007pointwise}.

The orientation estimation system 
    has shown a competitive yaw drift of $5^{\circ}$ per hour, given a low-cost sensor.

The visual-inertial position estimation is a Kalman Filter,
  with a state consisting of translation and linear velocity as in Eq.~\ref{eq:state}.

\begin{equation}
  \mu_{t|t-1} = \begin{bmatrix}
            t_x,
            v_x,
            t_y,
            v_y,
            t_z,
            v_z
         \end{bmatrix}^T
\label{eq:state}
\end{equation}

  The prediction step in Eq.~\ref{eq:predict} $(\mu, \Sigma)_{t-1|t-1} \rightarrow (\mu, \Sigma)_{t|t-1}$  runs at $500$ Hz on accelerometer measurements $a_t$, which have:
  \begin{enumerate*}[series = tobecont, itemjoin = \quad, label=\arabic*.]
      \item Been brought into the world coordinate frame by an orientation estimate;
      \item The gravity vector substracted;
      \item The noise filtered by a high-pass filter
  \end{enumerate*}.

The filter operates in 3D, but without a loss of generality, Eqs.~\ref{eq:predict}, \ref{eq:correct} are formulated in 1D for compactness.
$F$ is a state transition matrix, matrix $G$ maps an acceleration measurement to a state, matrix $H$ maps a state to a velocity measurement.

\begin{equation}
\begin{split}
  &\mu_{t|t-1} = F \mu_{t-1|t-1} + G a_t\\
  &\Sigma_{t|t-1} =  F\Sigma_{t-1|t-1}F^T + GQ_tG^T\\
  &F  = \begin{bmatrix}
        1 & \Delta t \\
        0 & 1
       \end{bmatrix} \quad
  G = \begin{bmatrix}
      \frac {{\Delta t}^2} {2} \\
      \Delta t
      \end{bmatrix} \quad
      \quad H = \begin{bmatrix} 0&1 \end{bmatrix}
\end{split}
\label{eq:predict}
\end{equation}

$Q_t$ and $R_t$ are accelerometer and velocity measurement covariances respectively,
  maintained as empirical distributions in a sliding window fashion.

The correction step in Eq.~\ref{eq:correct} $(\mu, \Sigma)_{t|t-1} \rightarrow (\mu, \Sigma)_{t|t}$ uses velocity $v_t$,
  supplied by a visual tracker or a motion capture system (the latter is the case in experiments, Sec.~\ref{sec:md_in_slam}).

\begin{align}
\begin{split}
  \mu_{t|t} &= \mu_{t|t-1} + K_t (v_t - H\mu_{t|t-1}) \\
  \Sigma_{t|t} &=  (I-K_tH)\Sigma_{t|t-1}\\
  K_t &= \Sigma_{t|t-1} H^T(H\Sigma_{t|t-1}H^T+R_t)^{-1}
\end{split}
\label{eq:correct}
\end{align}

Timestamps in state subscripts imply that the visual tracker and the IMU run by the aligned clock,
  which is not true in practice and moreover, the latency of visual updates varies.
The interpolation by agent kinematics would benefit the system accuracy
  and may be implemented by e.g. a sliding window Kalman Filter,
  but its practical realization is out of scope for the present study.

The main proposition is the way acceleration and velocity is provided to the Kalman Filter:
    $a_t$, $v_t$ are replaced with zero pseudo-updates $0^{3 \times 1}$
    when the system is in a \textit{stilless} mode and processed normally otherwise.
The mode transitioning is further explained.

\section{Motion Detector} \label{sec:motion_detector}
Pluto navigation system transitions to the pseudo-update mode by virtue of the motion detector,
which is a deep neural network-based two-class classifier,
   taking a 3D acceleration data stream as an input and outputting a label \{\textit{motion}, \textit{stillness}\}.

Windows of 3D acceleration data, registered during participants walking forward, backward, or sideways, regadless of a head rotating, tilting, or being kept still to the body are considered
\textit{motion}. Other windows of 3D acceleration data, when a participant stays still on feet, even if their head rotates or tilts, are considered \textit{stillness}.

The input and output tensors are $[\text{B}, 100, 3]$ and $[\text{B}, 1, 2]$ respectively
  with dimensions encoding a batch size, time domain, and data dimensionality.
During inference the batch size is $1$ and data are supplied by the sliding-window cache,
  which shifts by $1$ timestamp at consecutive inference calls.

TCN~(Temporal Convolutional Network), based on dilated causal convolutions,
  has demonstrated superior performance on a wide range of sequence modeling tasks~\cite{bai1803empirical}.
Experiments (Tab.~\ref{tab:nn_train}) show that TCN overperforms baselines accuracy by a notable
   margin in the accelerometer data classification.

\begin{figure}[h!]
\centering
\includegraphics[width=0.47\textwidth]{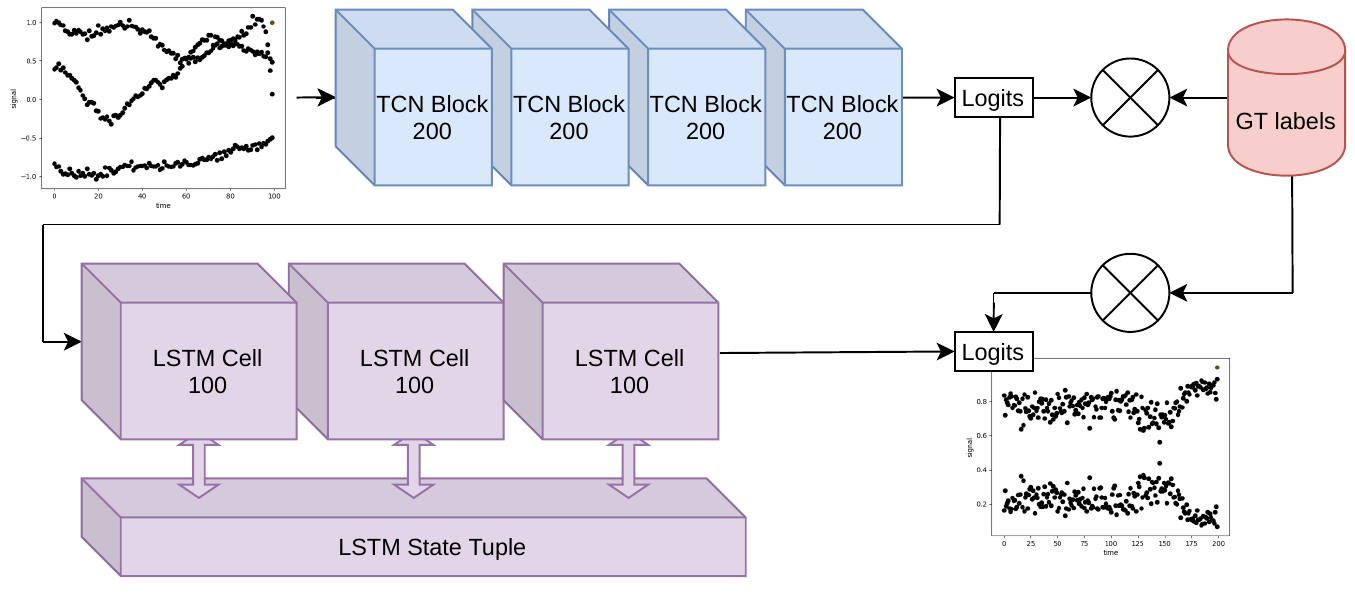}
\caption{Pluto motion detector architecture consists of the TCN classifier~(blue) and
         LSTM~(purple) for smoothing over the logits.
  The numbers after block names are hidden dimension sizes.
  The TCN blocks amount and kernel size were selected to cover all timestamps in a sliding-window.
  The input is a window of 3D acceleration data stream, outputs are likelihoods of possible agent states: \{\textit{motion}, \textit{stillness}\}.
  }
\label{gr:arch}
\end{figure}

Despite the higher accuracy, TCN is at a disadvantage from standpoint of false-positive rate,
  which is required to be as low as possible for a virtual reality application.
We attribute that to the TCN output being independent of the network inferences on previous timestamps.
The proposed solution~(Fig.~\ref{gr:arch}) stacks TCN with a stateful LSTM, which is fed with TCN logit outputs, hence
 explicitly taking the temporal context into account. Networks are trained separately.

\section{Experiments} \label{sec:experiments}
A thorough evaluation of Pluto navigation system comprises two parts:
\begin{enumerate*}[series = tobecont, itemjoin = \quad, label=\arabic*.]
\item Pluto motion detector classification performance is studied in comparison with several baselines;
\item Pluto navigation system positioning drift is evaluated under conditions of repeated tracking failures
\end{enumerate*}.
We proceed with the specification of collected data,
  which were made available for the benefit of the research community\footnote{\href{https://github.com/wf34/pluto}{https://github.com/wf34/pluto}}.

\subsection{Dataset}\label{sec:dataset}

The dataset was taken with a headset prototype, spanning 30 min in time, where 4 subjects, varying in age and gender
  were asked to explore a virtual reality scene, while moving freely and naturally (Fig.~\ref{gr:pluto}.a).
During the procudure following were collected:
\begin{enumerate*}[series = tobecont, itemjoin = \quad, label=\arabic*.]
\item IMU data (accelerometer, gyroscope, magmetometer);
\item 6D positioning ground truth
\end{enumerate*}.
High-precision 6D positioning ground truth data were collected with a motion capture system OptiTrack\textsuperscript{\textregistered} at 125~Hz.
The environment for the study was a well lighted lab room, free of dynamic agents other than a subject.
Recorded trajectories are rich with sporadic movements, side- and backward steps,
  participants lean and change direction and orientation restlessly, rotate and tilt their heads, while observing a virtual reality scene.
Data, registered from one of subjects, are held out for testing, remaining data are used in a network training.
Such a partitioning enables to verify the networks ability to generalize to inputs produced by a previously unseen person.

Experiments with the integrated navigation system (Sec.~\ref{sec:md_in_slam}) were carried out on a testing partition,
  which was split onto $16$ sequences of variable length ($5 \dots 12$ s) at random.
Each of these tracks is having a simulated tracking failure introduced, lasting a random time ($2 \dots 8$ s) at a random offset from the start.
The four tracks shown in Fig.~\ref{gr:navigation_bviews} are segments of a full sequence from Fig.~\ref{gr:pluto}.d.

\subsection{Motion Detection Performance}\label{sec:md_perf}
\begin{figure}[t!]
\centering
\includegraphics[width=.9\linewidth]{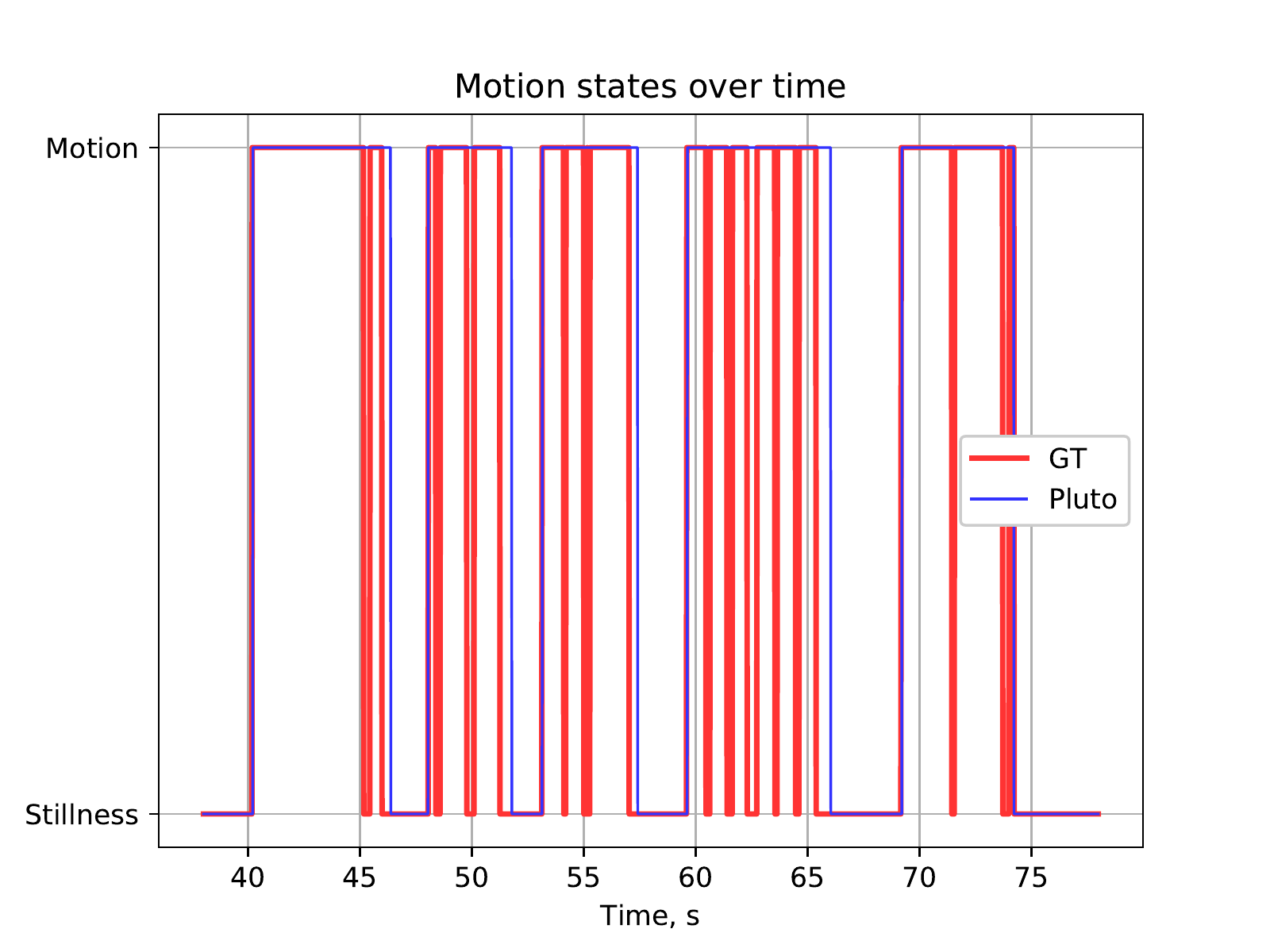}\\
\caption{Pluto motion detector output labels
         in comparison with motion capture ground truth labels (GT). A low false positive rate with low detection delays are demonstrated.}
\label{gr:labels}
\end{figure}

The agent state classification evaluation starts with a qualitative visualization of motion starts and stops,
as were registered by motion capture and classified by Pluto motion detector on a testing sequence in Fig.~\ref{gr:labels}.
Inferred labels do not lag and demonstrate a low false-positive rate.
One nuisance is that the system was unable to detect
    short stops between sequences of steps in $53 \dots 55$, $60 \dots 65$ s., due to the participants head still slightly moving,
    while the stepping paused. 
These micro stops are included in ground truth class labels,
  which were automatically converted from 6D ground truth trajectories with the following thresholding on velocity:
  $\Vert v \Vert_2 < 0.2\ \frac {\text{m}}{\text{s}}$.

\begin{figure}[t!]
\centering
\begin{overpic}[width=.9\linewidth]{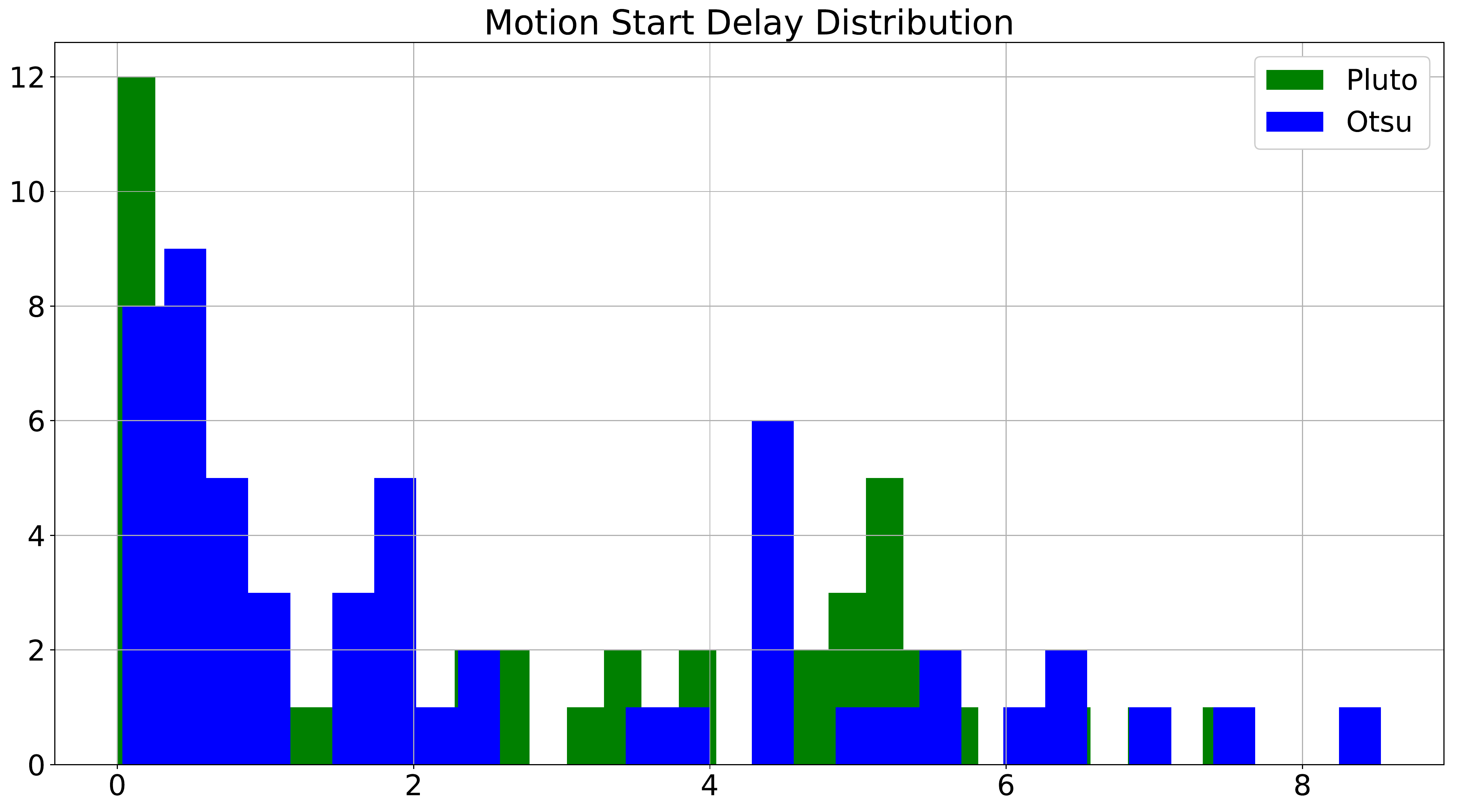}
  \put (-3, 54) {\scriptsize{Events, \#}}
  \put (45,-3) {\scriptsize{Time, s}}
\end{overpic}
\caption{Histogram of delays in detecting a motion start, [sec]. Pluto motion detector shows more low-delay events when compared to adaptive Otsu thresholding.
         Outliers in a right part of the domain were addressed in Sec.~\ref{sec:md_perf}}
\label{gr:mdd_distribs}
\end{figure}

Networks converge despite a few mislabelings, but labeling imperfections also produce outliers in evaluation.
Fig.~\ref{gr:mdd_distribs} shows detection delay histograms for two most competitive algorithms.
Classification time series from Fig.~\ref{gr:labels} allow to claim that longer delays~($>5$ s)
are absent in reality and appear on a figure due to ground truth event mislabelings.

Pluto motion detector is compared to LSTM, CNN, TCN, and a variant~\cite{pluto_patent}
  of signal processing classic, Otsu thresholding~\cite{otsu1979threshold}.
Results in Tab.~\ref{tab:nn_train} suggest that TCN, a core building block of Pluto motion detector, is superior in terms of accuracy, training time and weights amount.
The variant of Adaptive Otsu thresholding, which was employed in an evaluation,
  operates on a 1D signal (a norm of registered acceleration $|| a ||_2$),
            maintains its histogram and updates a threshold, which splits
            the histogram into two parts with a maximal inter-class variance.
A relation between a current sample and the threshold value is used for classification.

\begin{table}[t!]
\caption{Neural Networks Training and Inference}
\centering{
\tiny{
\setlength{\tabcolsep}{0.3em}
\sisetup{detect-weight=true,detect-inline-weight=math}
\begin{tabular}{l|c|c|c|cc|cc|cc}
\toprule
  Model & \#weights, M & \vtop{\hbox{\strut Depth}\hbox{\strut (in blocks)}} & \vtop{\hbox{\strut Train time}\hbox{\strut (per epoch, h)}}  & \multicolumn{2}{c}{Train} & \multicolumn{2}{c}{Validation} & \multicolumn{2}{c}{Test} \\
   &  &  &  & accuracy, \% & loss & accuracy, \% & loss & accuracy, \% & loss \\
\midrule
  CNN & 2.7 & 5 & \textbf{0.28} & 0.899 & 0.324 & 0.751 & 0.702 & 0.723 & 1.013 \\
  LSTM & 1.9 & 4 & 1.43 & 0.853 & 0.473 & 0.829 & 0.596 & 0.820 & 0.687 \\
  TCN & \textbf{1.4} & 4 & 0.62 & \textbf{0.948} & \textbf{0.139} & \textbf{0.889} & \textbf{0.304} & \textbf{0.871} & \textbf{0.371} \\
  Pluto & 1.4 + 0.04 & 4+3 & 1.55 & 0.901 & 0.244 & 0.811 & 0.437 & 0.869 & 0.376 \\
\bottomrule
\end{tabular}
}
}
\label{tab:nn_train}\\
\vspace{0.4em}
\scriptsize{TCN~(a core building block of Pluto) is the best architecture for signal processing of accelerometer}.
Pluto motion detector, which connects TCN to LSTM, trades off a moderate accuracy loss for a radical false positive rate reduction~(Tab.~\ref{tab:mdd_mfpi}).
\end{table}

All evaluated networks were trained with a 3D signal,
  which is a registered acceleration in the global coordinate frame.
TCN and CNN networks were trained with Adam optimizer, LSTM with RMSProp and the cross-entropy loss was used for all.
The hyperparameter choice (the sliding window size of $100$ timestamps = $0.2$~s) was done by a grid search (with boundaries $0.05 \dots 3$~s) on a training partition with the baseline CNN network.
All networks were implemented in Tensorflow,
underwent several iterations of coarse-to-fine tuning, have approximately the same size and were trained for the same time.
The times per epoch in Tab.~\ref{tab:nn_train} were registered for CPU training at $20$ cores on Intel Xeon E5.
Pluto motion detector inference is tangible for realtime running on the headset prototype at a decreased frequency~(10 Hz).

The achieved accuracy of 87\%~(Tab.~\ref{tab:md_metrics}) could have been improved by the dataset increase, due to
 movements of the test subject (and their IMU accelerations) laying outside of a network domain, because only  a small set of anthropometrically different study subjects comprise the training partition.
Nonetheless, network overfitting was avoided, which evidences from accuracy being approximately equal on training and testing dataset partitions.

Attempted network training approaches, which have shown no classification improvement are:
  data augmentation by affine transformations, input dimensionality extension by gyroscope and magnetic data,
  pretraining on synthetic\footnote{\href{https://github.com/Aceinna/gnss-ins-sim}{https://github.com/Aceinna/gnss-ins-sim}} accelerations.

\begin{table}[t!]
\caption{Motion Detection Metrics}
\centering{
\scriptsize{
\setlength{\tabcolsep}{0.3em}
\begin{tabular}{l|c c c c|}
\toprule
  Method & Accuracy & Precision & Recall & $F_1$ \\
\midrule
  Otsu   & 0.746    & 0.758     & 0.745  & 0.752 \\
  \textbf{Pluto}  & \textbf{0.874}    & \textbf{0.825}     & \textbf{0.960}  & \textbf{0.887} \\
\bottomrule
\end{tabular}\\
\label{tab:md_metrics}
\vspace{0.4em}
  Pluto motion detector shows a superior classification performance to the baseline on a dataset testing partition~(Sec.~\ref{sec:md_perf}).
}
}
\end{table}

User experience tests in virtual reality have shown a low correlation with the accuracy metric because the temporal context is not captured:
  the metric value is the same, whether a detection is off just by a one timestamp or $50$.
To adjust for that, an additional performance evaluation was conducted with~\cite{basseville1993detection}:
\begin{enumerate*}[series = tobecont, itemjoin = \quad, label=\arabic*.]
  \item Mean detection delay;
  \item Mean interval between false-positive detections
\end{enumerate*}.
These metrics, contrary to accuracy, precision, recall, $F_1$ are in a temporal domain.
The detection delay is a time in seconds since a labeled ground truth event until its true positive detection, issued by the detector.
The interval between false positive detections is a time in seconds between consecutive detections, issued by the detector and not having labeled ground truth events corresponding to them.

Due to Tab.~\ref{tab:mdd_mfpi}, Pluto motion detector issues false positive detections 4 times less often than Otsu on average, while has approximately same true positive detection delays.
To capture results beyond a mean and variance, the histogram of motion start detection delays is shown in Fig.~\ref{gr:mdd_distribs}.

A sizable variance in metric magnitudes is found in Tab.~\ref{tab:mdd_mfpi}. It is due to the fact that neural network sensitivity is hard to adjust: 
LSTM has converged to be very sensitive to signal changes while producing more false positives, while TCN is less prone to false positive detections, but results in longer delays.
The proposed motion detector Pluto combines advantages of both, enabling the higher responsiveness with fewer false positives.

\begin{table}[t!]
\caption{Motion Detection Performance, [sec]}
\centering{
\scriptsize{
\setlength{\tabcolsep}{0.3em}
\begin{tabular}{l|cc|cc|cc|cc|}
   \toprule
    \multirow{2}{*}{Method} & \multicolumn{4}{c}{Starts Detection} & \multicolumn{4}{c}{Stops Detection} \\
                            & \multicolumn{2}{c}{Detection Delay$^{\textit{l}}$} & \multicolumn{2}{c}{\vtop{\hbox{\strut Interval}\hbox{\strut between}\hbox{\strut false positives$^{\textit{h}}$}}}
                            & \multicolumn{2}{c}{Detection Delay$^{\textit{l}}$} & \multicolumn{2}{c}{\vtop{\hbox{\strut Interval}\hbox{\strut between}\hbox{\strut false positives$^{\textit{h}}$}}} \\
    & $\mu$ & $\sigma$ & $\mu$ & $\sigma$ & $\mu$ & $\sigma$ & $\mu$ & $\sigma$ \\
    \midrule
     Otsu  & 2.409          &  2.324 &  8.856          & 12.013 & 1.435          & 1.205 & 11.060          & 10.829 \\
     CNN   & 2.634          & 17.182 &  0.487          &  0.890 & 0.316          & 0.568 &  0.492          &  0.881 \\
     LSTM  & \textbf{0.292} &  0.346 &  0.188          &  0.431 & \textbf{0.252} & 0.246 &  0.188          &  0.429 \\
     TCN  & 1.893          &  1.935 &  2.125          &  2.969 & 1.071          & 1.195 &  2.160          &  3.242 \\
     Pluto & 2.389          &  2.128 & \textbf{14.889} & 18.425 & 1.771          & 1.427 & \textbf{40.093} & 48.488 \\
    \bottomrule
\end{tabular}\\
\label{tab:mdd_mfpi}
\vspace{0.4em}
  Pluto motion detector shows low detection delays and high intervals between false alarms,
  outperforming competitors in $\frac {\text{MDD}} {\text{MFPI}}$ for start and stop events both.
  $^{\textit{l}}$ means lower is better; $^{\textit{h}}$ means higher is better.
}
}
\end{table}

\subsection{Effects of Motion Detection on Navigation} \label{sec:md_in_slam}
The motion detector capable of discerning an agents stillness and motion
  may benefit a navigation system, as proposed in Sec.~\ref{sec:overview}.
The claim is evaluated in simulated experiments, conducted on $16$ sequences, taken from testing data, as described in Sec.~\ref{sec:dataset}.
The navigation system is a Kalman filter, with prediction steps by IMU acceleration and correction steps by linear velocity.
Velocity and orientation estimates are provided by a motion capture system.
Velocity updates are not provided during a simulated tracking failure.

\begin{figure}[t!]
\centering
\begin{tabular}{cc}
  \includegraphics[width=.47\linewidth]{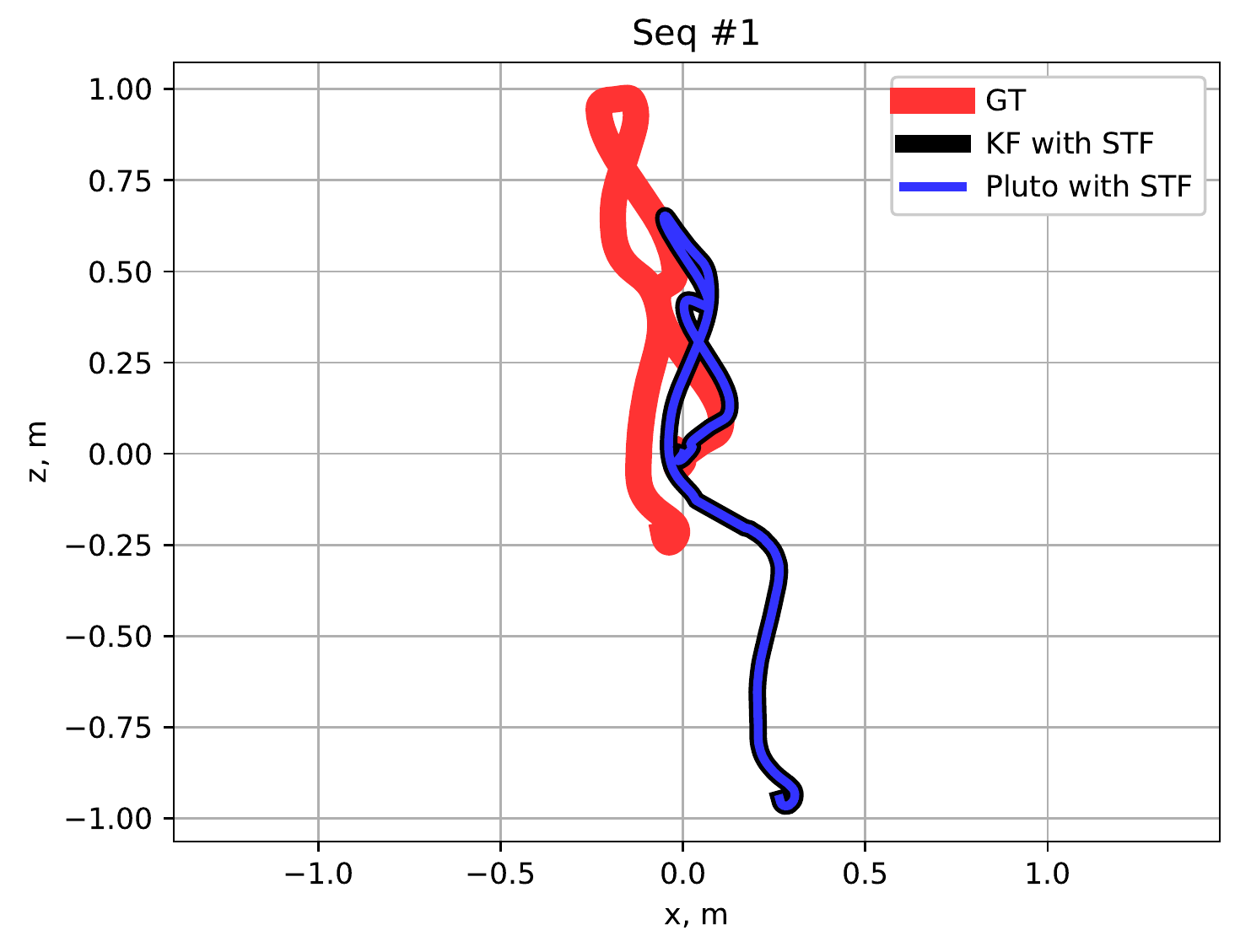} &
  \includegraphics[width=.47\linewidth]{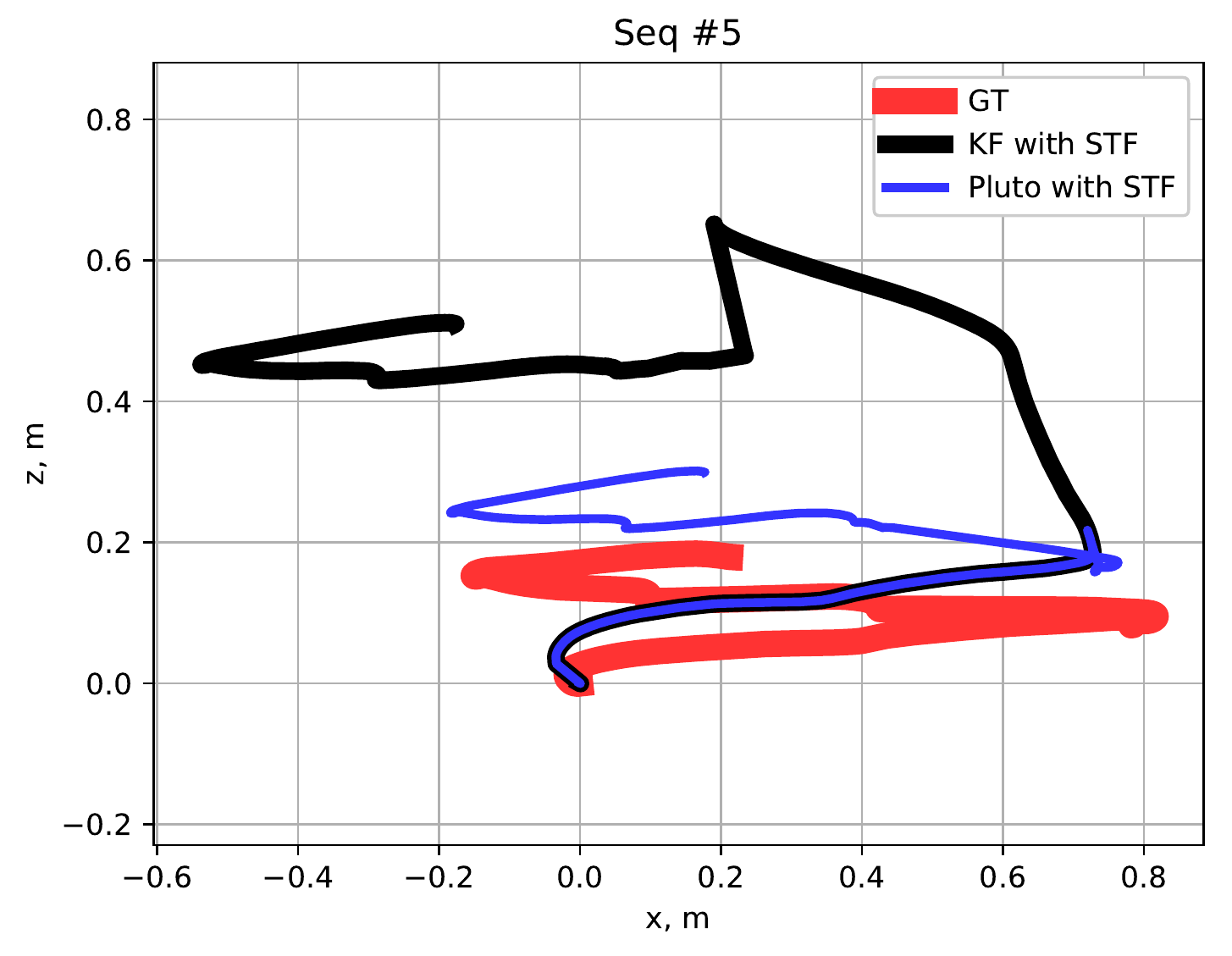} \\
  \includegraphics[width=.47\linewidth]{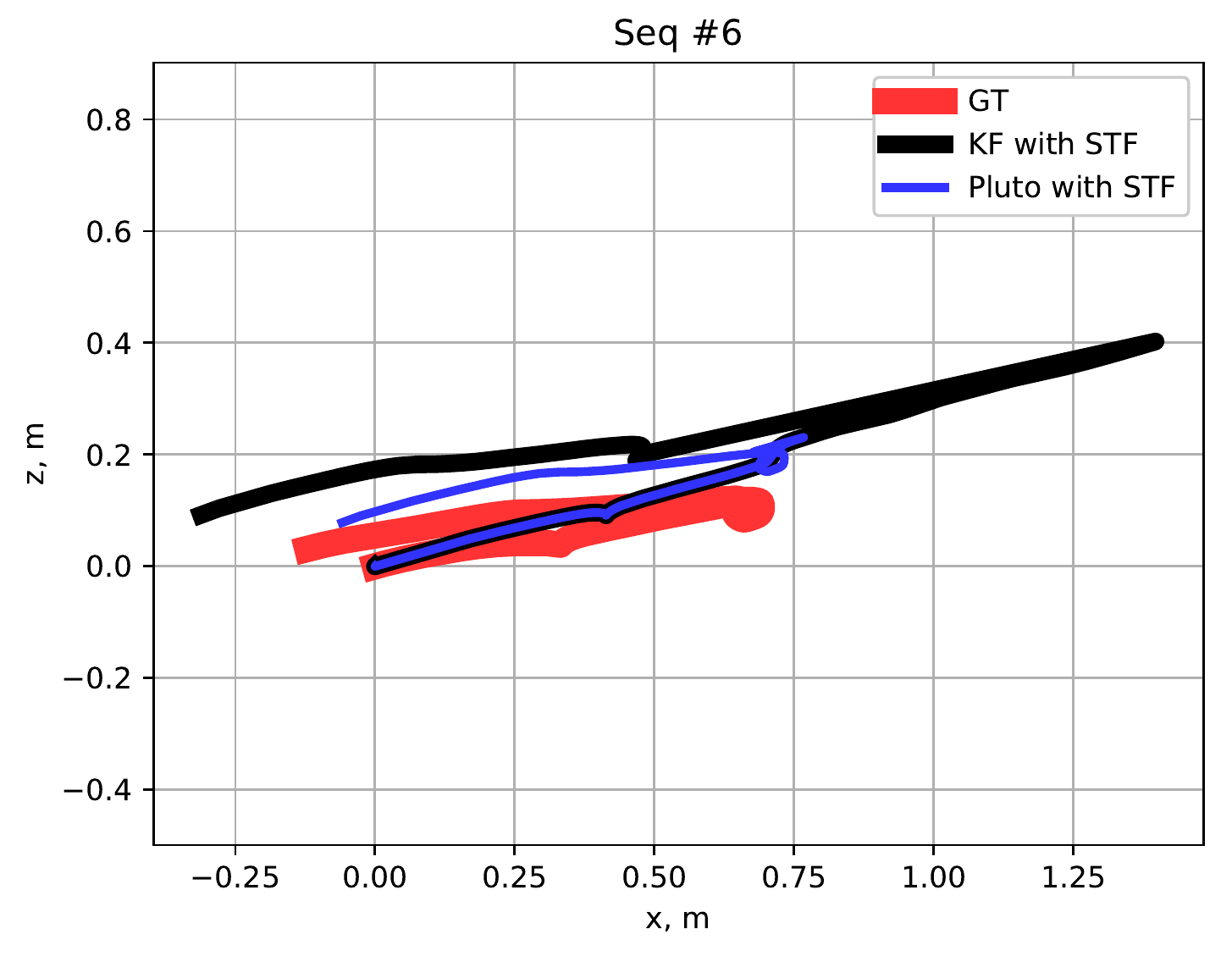} &
  \includegraphics[width=.47\linewidth]{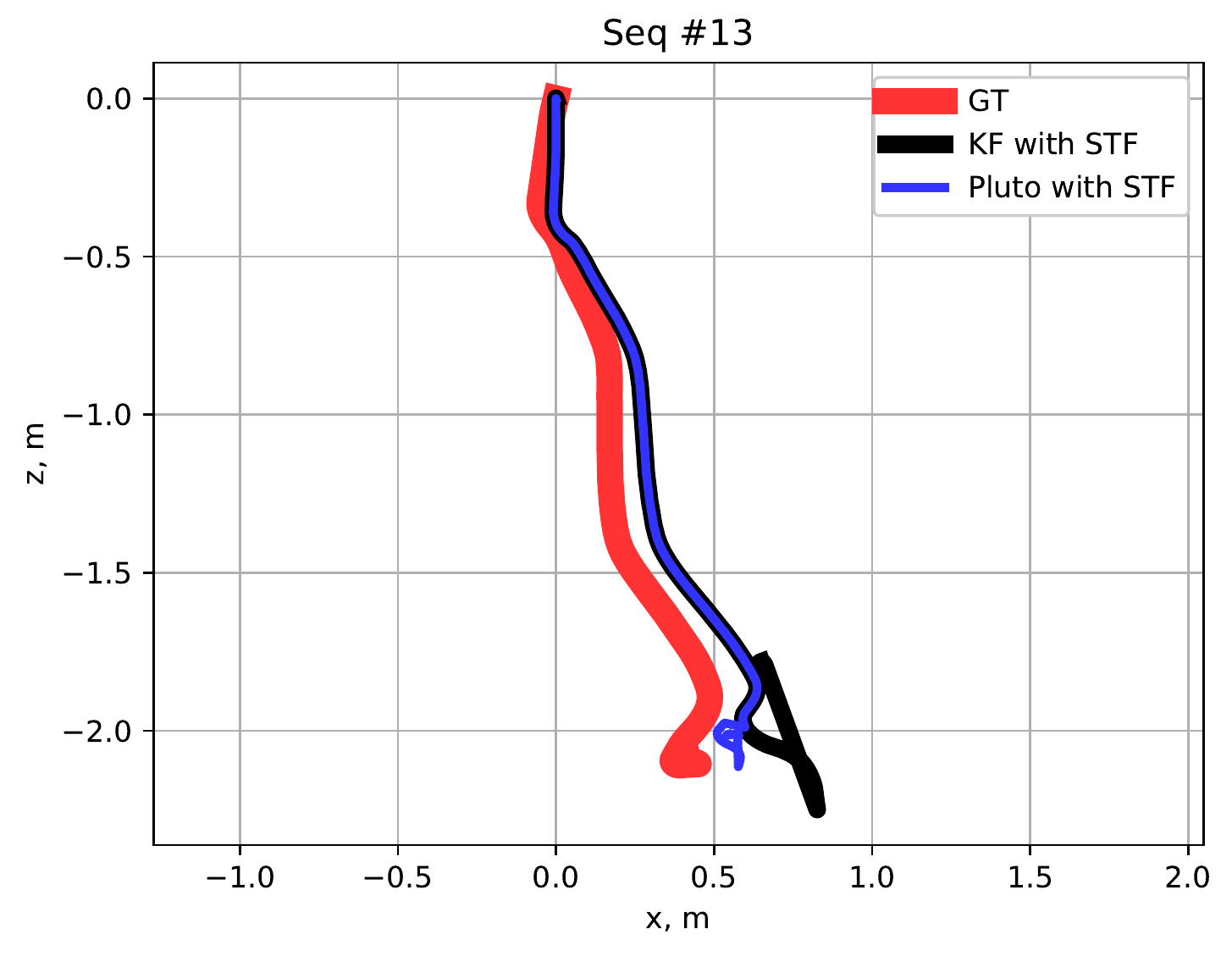} \\
\end{tabular}
\caption{
  Navagation system trajectories, produced in Sec.~\ref{sec:md_in_slam} experiments and selected at random, provide
  comparison of Pluto~(blue) with the ground truth~(red) and the baseline~(Kalman filter, black).
  Pluto navagation system is less prone to drift during tracking failures.
  }
\label{gr:navigation_bviews}
\end{figure}

The navigation precision is estimated by Relative Positioning Drift, which is a fraction of error accumulated by a navigation system since the specified point in time.
In a present case, the reference timestamp was chosen such that a relative path between the reference timestamp and the current timestamp is $1$ m,
  measured by a motion capture system.
The mean average relative drift over $16$ sequences and its variance are presented in Tab.~\ref{tab:rel_drift}.
Also, four sequences were picked at random and visualized in Fig.~\ref{gr:navigation_bviews}.

Obtained results show that motion detector capabilities may substantially improve the resilience of a navigation system in a face of tracking failures.

\begin{table}[t!]
\caption{Relative Positioning Drift, \%}
\centering{
\scriptsize{
\begin{tabular}{l|cc|cc|cc}
   \toprule
    \multirow{2}{*}{} & \multicolumn{2}{c}{if no STF KF = Pluto} & \multicolumn{2}{c}{Pluto with STF} & \multicolumn{2}{c}{KF with STF}  \\
    & $\mu$ & $\sigma$ & $\mu$ & $\sigma$ & $\mu$ & $\sigma$ \\
    \midrule
     Avg over $16$ seqs & 2.52 & 0.93 & 7.58 & 7.28 & 12.95 & 15.07 \\
    \bottomrule
\end{tabular}
}
}
\label{tab:rel_drift}\\
\vspace{0.4em}
\scriptsize{The baseline (KF) shows 5 times drift increase under conditions of simulated tracking failures~(STF)
    relative to a failure-free tracking, while drift of the proposed Pluto system grows moderately.}
\end{table}

\section{Conclusions} \label{sec:conclusions}
A neural network-based motion detector was proposed,
  which classifies \textit{motion} and \textit{stillness} states of an agent with 87\% accuracy
      while generalizing to the motion pattern of a person, who was not present in training data.
The detector is capable of online operation,
  with the mean interval between false-positive detections more than 14 seconds,
    which is longer than a typical relocalization time of a computer vision-based navigation system.

The impact of the motion classification on navigation was demonstrated,
  decreasing positioning drift by 40\% in a repeated tracking failure scenario.
A practical drift reduction could be even higher because virtual reality users tend to move less during a tracking failure occurrence.
In that case, the motion detector would register stillness and update a system kinematic state accordingly.
A working prototype was developed and tested;
  a high-quality dataset with motion capture ground truth
  was obtained and made available.

\thanks{We are grateful to Ilya Nedelko, Sergiy Pometun, Oleg Muratov, Tarek Dakhran,
 Andriy Marchenko and Mikhail Rychagov for their contributions and support.
We extend gratitude to anonymous reviewers for the thoughtful input.
}

\printbibliography

@inproceedings{lavalle2014head,
  title={Head tracking for the Oculus Rift},
  author={LaValle, Steven M and Yershova, Anna and Katsev, Max and Antonov, Michael},
  booktitle={International Conference on Robotics and Automation (ICRA)},
  pages={187--194},
  year={2014},
  organization={IEEE}
}

@inproceedings{gusenbauer2010self,
  title={Self-contained indoor positioning on off-the-shelf mobile devices},
  author={Gusenbauer, Dominik and Isert, Carsten and Kr{\"o}sche, Jens},
  booktitle={Indoor positioning and indoor navigation (IPIN)},
  pages={1--9},
  year={2010},
  organization={IEEE}
}

@article{puyol2014pedestrian,
  title={Pedestrian simultaneous localization and mapping in multistory buildings using inertial sensors},
  author={Puyol, Maria Garcia and Bobkov, Dmytro and Robertson, Patrick and Jost, Thomas},
  journal={IEEE Transactions on Intelligent Transportation Systems},
  volume={15},
  number={4},
  pages={1714--1727},
  year={2014},
  publisher={IEEE}
}

@article{harle2013survey,
  title={A survey of indoor inertial positioning systems for pedestrians.},
  author={Harle, Robert},
  journal={IEEE Communications Surveys and Tutorials},
  volume={15},
  number={3},
  pages={1281--1293},
  year={2013}
}

@article{sun2014zupt,
  author = {Sun, X. and Wu, K. and Li, Y. and Di, Kaichang},
  year = {2014},
  month = {03},
  pages = {239},
  title = {A Zupt-Based Method for Astronaut Navigation on Planetary Surface and Performance Evaluation under Different Locomotion Patterns},
  volume = {XL-4},
  journal = {ISPRS - International Archives of the Photogrammetry, Remote Sensing and Spatial Information Sciences},
  doi = {10.5194/isprsarchives-XL-4-239-2014}
}

@article{feliz2009pedestrian,
  author = {Feliz, Raul and Zalama, Eduardo and Gómez-García-Bermejo, Jaime},
  year = {2009},
  month = {01},
  pages = {},
  title = {Pedestrian Tracking Using Inertial Sensors},
  volume = {3},
  journal = {Journal of Physical Agents},
  doi = {10.14198/JoPha.2009.3.1.05}
}

@inproceedings{goyal2011strap,
  title={Strap-down pedestrian dead-reckoning system},
  author={Goyal, Pragun and Ribeiro, Vinay J and Saran, Huzur and Kumar, Anshul},
  booktitle={Indoor Positioning and Indoor Navigation (IPIN)},
  pages={1--7},
  year={2011},
  organization={IEEE}
}

@article{sun2008activity,
author = {Sun, Zuolei and Mao, Xuchu and Tian, Weifeng and Zhang, Xiangfen},
year = {2008},
month = {11},
pages = {015203},
title = {Activity Classification and Dead Reckoning for Pedestrian Navigation with Wearable Sensors},
volume = {20},
journal = {Measurement Science and Technology},
doi = {10.1088/0957-0233/20/1/015203}
}

@inproceedings{skog2010evaluation,
  title={Evaluation of zero-velocity detectors for foot-mounted inertial navigation systems},
  author={Skog, Isaac and Nilsson, John-Olof and H{\"a}ndel, Peter},
  booktitle={Indoor Positioning and Indoor Navigation (IPIN)},
  pages={1--6},
  year={2010},
  organization={IEEE}
}

@inproceedings{xiaofang2014applications,
  title={Applications of zero-velocity detector and Kalman filter in zero velocity update for inertial navigation system},
  author={Xiaofang, Li and Yuliang, Mao and Ling, Xie and Jiabin, Chen and Chunlei, Song},
  booktitle={Guidance, Navigation and Control Conference (CGNCC)},
  pages={1760--1763},
  year={2014},
  organization={IEEE}
}

@inproceedings{beauregard2006helmet,
  title={A helmet-mounted pedestrian dead reckoning system},
  author={Beauregard, St{\'e}phane},
  booktitle={3rd International Forum on Applied Wearable Computing (IFAWC)},
  pages={1--11},
  year={2006},
  organization={VDE}
}

@inproceedings{gjoreski2015recognizing,
  title={Recognizing atomic activities with wrist-worn accelerometer using machine learning},
  author={Gjoreski, Martin and Gjoreski, Hristijan and Lu{\v{s}}trek, Mitja and Gams, Matja{\v{z}}},
  booktitle={Proceedings of the 18th International Multiconference Information Society (IS), Ljubljana, Slovenia},
  pages={10--11},
  year={2015}
}

@article{ordonez2016deep,
  title={Deep convolutional and lstm recurrent neural networks for multimodal wearable activity recognition},
  author={Ord{\'o}{\~n}ez, Francisco Javier and Roggen, Daniel},
  journal={Sensors},
  volume={16},
  number={1},
  pages={115},
  year={2016},
  publisher={Multidisciplinary Digital Publishing Institute}
}

@inproceedings{anguita2012human,
author = {Anguita, Davide and Ghio, Alessandro and Oneto, Luca and Parra, Xavier and Reyes-Ortiz, Jorge},
year = {2012},
month = {12},
pages = {216-223},
title = {Human Activity Recognition on Smartphones Using a Multiclass Hardware-Friendly Support Vector Machine},
volume = {7657},
isbn = {978-3-642-35394-9},
journal = {Ambient Assist. Living Home Care},
doi = {10.1007/978-3-642-35395-6_30}
}

@article{ozyagcilar2012calibrating,
  title={Calibrating an ecompass in the presence of hard and soft-iron interference},
  author={Ozyagcilar, Talat},
  journal={Freescale Semiconductor Ltd},
  pages={1--17},
  year={2012}
}

@article{foi2007pointwise,
  title={Pointwise shape-adaptive DCT for high-quality denoising and deblocking of grayscale and color images},
  author={Foi, Alessandro and Katkovnik, Vladimir and Egiazarian, Karen},
  journal={IEEE Transactions on Image Processing},
  volume={16},
  number={5},
  pages={1395--1411},
  year={2007},
  publisher={IEEE}
}

@book{basseville1993detection,
  title={Detection of abrupt changes: theory and application},
  publisher={Prentice Hall Englewood Cliffs}
  author = {Basseville, Michèle and Nikiforov, Igor},
  year = {1993},
  month = {04},
  pages = {},
  title = {Detection of Abrupt Change Theory and Application},
  volume = {15},
  isbn = {0-13-126780-9}
}

@article{otsu1979threshold,
  title={A threshold selection method from gray-level histograms},
  author={Otsu, Nobuyuki},
  journal={IEEE transactions on systems, man, and cybernetics},
  volume={9},
  number={1},
  pages={62--66},
  year={1979},
  publisher={IEEE}
}

@online{chevalier,
  title = {LSTMs for Human Activity Recognition},
  url = {https://github.com/guillaume-chevalier/LSTM-Human-Activity-Recognition},
  year = {2018},
  urldate = {2018-05-31}
}

@article{quinones2018hive,
  title={HIVE Tracker: a tiny, low-cost, and scalable device for sub-millimetric 3D positioning},
  author={Colomer, Darío R. and Lopes, Gonçalo and Kim, Danbee and Honnet, Cedric and Moratal, David and Kampff, Adam},
  journal={Augmented Human},
  volume={9},
  year={2018},
  pages = {1-8},
}

@article{jones2011visual,
  title={Visual-inertial navigation, mapping and localization: A scalable real-time causal approach},
  author={Jones, Eagle S and Soatto, Stefano},
  journal={The International Journal of Robotics Research},
  volume={30},
  number={4},
  pages={407--430},
  year={2011},
  publisher={SAGE Publications Sage UK: London, England}
}

@inproceedings{jiang2015human,
  title={Human tracking using wearable sensors in the pocket},
  author={Jiang, Wenchao and Yin, Zhaozheng},
  booktitle={IEEE Global Conference on Signal and Information Processing (GlobalSIP)},
  pages={958--962},
  year={2015},
  organization={IEEE}
}

@article{verplaetse1996inertial,
  title={Inertial proproceptive devices: Self-motion-sensing toys and tools},
  author={Verplaetse, Christopher},
  journal={IBM Systems Journal},
  volume={35},
  number={3.4},
  pages={639--650},
  year={1996},
  publisher={IBM}
}

@inproceedings{yan2018ridi,
  title={RIDI: Robust IMU double integration},
  author={Yan, Hang and Shan, Qi and Furukawa, Yasutaka},
  booktitle={Proceedings of the European Conference on Computer Vision (ECCV)},
  pages={621--636},
  year={2018}
}

@inproceedings{windau2016walking,
  title={Walking compass with head-mounted IMU sensor},
  author={Windau, Jens and Itti, Laurent},
  booktitle={2016 IEEE International Conference on Robotics and Automation (ICRA)},
  pages={5542--5547},
  year={2016},
  organization={IEEE}
}

@inproceedings{rantanen2018,
  author = {Rantanen, Jesperi and Makela, Maija and Ruotsalainen, Laura and Kirkko-Jaakkola, Martti},
  year = {2018},
  month = {09},
  pages = {206-212},
  title = {Motion Context Adaptive Fusion of Inertial and Visual Pedestrian Navigation},
  doi = {10.1109/IPIN.2018.8533872}
}

@inproceedings{wagstaff2018lstm,
  title={LSTM-based zero-velocity detection for robust inertial navigation},
  author={Wagstaff, Brandon and Kelly, Jonathan},
  booktitle={2018 International Conference on Indoor Positioning and Indoor Navigation (IPIN)},
  pages={1--8},
  year={2018},
  organization={IEEE}
}

@inproceedings{feigl2020rnn,
 title={RNN-aided human velocity estimation from a single IMU},
 author={Feigl, Tobias and Kram, Sebastian and Woller, Philipp and Siddiqui, Ramiz H and Philippsen, Michael and Mutschler, Christopher},
 journal={Sensors},
 volume={20},
 number={13},
 pages={3656},
 year={2020},
 publisher={Multidisciplinary Digital Publishing Institute}
}

@inproceedings{cortes2018deep,
  title={Deep Learning Based Speed Estimation for Constraining Strapdown Inertial Navigation on Smartphones},
  author={Cort{\'e}s, Santiago and Solin, Arno and Kannala, Juho},
  booktitle={IEEE 28th International Workshop on Machine Learning for Signal Processing (MLSP)},
  pages={1--6},
  year={2018},
  organization={IEEE}
}

@article{bai1803empirical,
  title={An empirical evaluation of generic convolutional and recurrent networks for sequence modeling. arXiv 2018},
  author={Bai, S and Kolter, JZ and Koltun, V},
  journal={arXiv preprint arXiv:1803.01271}
}

@article{silva2019end,
  title={End-to-End Learning Framework for IMU-Based 6-DOF Odometry},
  author={Silva do Monte Lima, Jo{\~a}o Paulo and Uchiyama, Hideaki and Taniguchi, Rin-ichiro},
  journal={Sensors},
  volume={19},
  number={17},
  pages={3777},
  year={2019},
  publisher={Multidisciplinary Digital Publishing Institute}
}

@misc{pluto_patent,
 title={Method and device for strap down inertial navigation},
 author={Migukin, A and Kovalenko, D and Ryabkova, S and Chernov, V},
 year={RU Patent 2685767C1, 2018.08.13}
}

@article{liu2020tlio,
  title={TLIO: Tight Learned Inertial Odometry},
  author={Liu, Wenxin and Caruso, David and Ilg, Eddy and Dong, Jing and Mourikis, Anastasios I and Daniilidis, Kostas and Kumar, Vijay and Engel, Jakob},
  journal={IEEE Robotics and Automation Letters},
  volume={5},
  number={4},
  pages={5653--5660},
  year={2020},
  publisher={IEEE}
}

\end{document}